\documentclass{article}[10pt]

\usepackage[english]{babel}
\usepackage{svg}

\usepackage[a4paper,top=2cm,bottom=2cm,left=2.5cm,right=2.5cm,marginparwidth=1.75cm]{geometry}

\usepackage{amsmath}
\usepackage{amssymb}
\usepackage{dsfont}
\usepackage{graphicx}
\usepackage[colorlinks=true, allcolors=blue]{hyperref}
\usepackage[explicit]{titlesec}
\usepackage{authblk}

\usepackage{tikz}
\usetikzlibrary{shapes.geometric, arrows}

\newcommand\eref[1]{Eq.(\ref{#1})}
\newcommand\fref[1]{Fig.~\ref{#1}}
\newcommand\sref[1]{Section.~\ref{#1}}
\newcommand\aref[1]{Appendix.~\ref{#1}}

\providecommand{\keywords}[1]
{
  \small	
  \textbf{\textit{Keywords---}} #1
}

\title{Deep tensor networks with matrix product operators}
\author{Bojan Žunkovič \footnote{bojan.zunkovic@fri.uni-lj.si}}
\affil{Faculty of Computer and Information science, University of Ljubljana, Ljubljana, Slovenia}

\date{}

\begin{document}
\thispagestyle{empty}
\maketitle

\begin{abstract}
We introduce deep tensor networks, which are exponentially wide neural networks based on the tensor network representation of the weight matrices. We evaluate the proposed method on the image classification (MNIST, FashionMNIST) and sequence prediction (cellular automata) tasks. In the image classification case, deep tensor networks improve our matrix product state baselines and achieve 0.49\% error rate on MNIST and 8.3\% error rate on FashionMNIST. In the sequence prediction case, we demonstrate an exponential improvement in the number of parameters compared to the one-layer tensor network methods. In both cases, we discuss the non-uniform and the uniform tensor network models and show that the latter generalizes well to different input sizes.
\end{abstract}


\section{Introduction}
Tensor networks are ubiquitous in the field of low-dimensional many-body quantum physics due to their efficient description of short-range correlations and the locality of prototypical quantum Hamiltonians. Motivated by the successful approximation of quantum amplitudes, tensor networks have also been applied to machine learning problems (e.g. image classification \cite{stoudenmire2016supervised,stoudenmire2018learning, efthymiou2019tensornetwork, liu2019machine,martyn2020entanglement,meng2020residual,chen2021residual,kong2021quantum}, generative modelling \cite{cheng2019tree,stokes2019probabilistic,sun2020generative,liu2021tensor}, sequence and language modelling \cite{pestun2017tensor,guo2018matrix, bradley2020modeling, bradley2020language}, anomaly detection \cite{wang2020anomaly,streit2020network}). Adopting tensor network methods in machine learning enables compression of neural networks \cite{novikov2015tensorizing,garipov2016ultimate,hrinchuk2019tensorized}, adaptive training algorithms \cite{reyes2021multi}, derivation of interesting generalisation bounds \cite{bradley2020modeling}, information theoretical insight \cite{cohen2016expressive,deng2017quantum,levine2017deep,glasser2019expressive}, and new connections between machine learning and physics \cite{chen2018equivalence,dymarsky2021tensor,adhikary2021quantum}. On the other hand, apart from a few exceptions (e.g. anomaly detection \cite{wang2020anomaly}), tensor networks do not surpass the performance of standard neural network approaches. In particular, the state-of-the-art tensor networks achieve lower accuracy on simple image datasets (MNIST, FashionMNIST) than the best or even benchmark neural networks \cite{mcdonnell2015enhanced}. Moreover, recent studies of correlations in several image datasets (MNIST \cite{cheng2018information,martyn2020entanglement,convy2021mutual,lu2021tensor}, FashionMNIST \cite{lu2021tensor}, CIFAR10 \cite{lu2021tensor}, and TinyImages \cite{convy2021mutual}), imply a limited utility of tensor networks (in particular matrix product states) for image classification and generation due to long-range correlations in image datasets.

Tensor networks that have been applied to machine learning problems include matrix product states (MPS)\cite{stoudenmire2016supervised,efthymiou2019tensornetwork}, multi-scale renormalisation ansatz \cite{stoudenmire2018learning,kong2021quantum,cong2019quantum}, tree tensor networks \cite{felser2021quantum}, projected entangled pair states (PEPS) \cite{martyn2020entanglement,cheng2021supervised}, and matrix product operators (MPO) \cite{guo2018matrix, guo2020tensor}. Besides tensor networks, several useful numerical techniques for their optimisation have been adopted and applied in the machine learning context. In the one dimensional case, a common optimisation algorithm is the density matrix renormalisation group (DMRG) which is analogous to the alternating least squares. The main computational tool of DMRG is the singular value decomposition (SVD). From the perspective of quantum simulations, the essential benefits of DMRG are its efficiency and precise control over the approximation error. However, it is only locally optimal and does not preserve conservation laws. A major generalisation of DMRG is the time-dependent variational principle on the matrix product state manifold (MPS-TDVP) \cite{haegeman2016unifying} which preserves conservation laws and is globally optimal. Importantly, the MPS-TDVP has the same asymptotic complexity as DMRG and does not leave the variational manifold, i.e. does not rely on the SVD to reduce the size of the matrix product state.

We will introduce an MPO layer analogous to the MPS-TDVP which does not increase the size of embeddings/inputs. This will enable us to stack many MPO layers in a deep tensor network which is exponentially more expressive than the shallow one-layer counterparts. Interestingly, we can interpret the MPO layer as a generalisation of the linear dot-attention mechanism, which considers also higher-order correlations. We will evaluate the introduced (deep) tensor networks on the image classification task (MNIST and Fashion MNIST) and the sequence prediction task (cellular automata).

In \sref{sec:tnml}, we review standard tensor network methods, before introducing the MPO layer based on the MPS-TDVP in Sections \ref{sec:mps-tdvp} and  \ref{sec:deeptn}. We discuss the relation between the MPO layers and standard neural networks in \sref{sec:tn-nn}. Finally, we evaluate the deep tensor networks on the image classification task in \sref{sec:MPS classification} and on the sequence prediction task in \sref{sec:sequence}. We conclude in \sref{sec:conclusion}.

\section{One-layer tensor networks}
\label{sec:tnml}
Before introducing deep tensor networks, we will resume the basics of the tensor-network image classification and sequence prediction problems \cite{stoudenmire2016supervised,guo2018matrix,efthymiou2019tensornetwork}. We assume that the input is a one-dimensional vector $\vec{x}$. In the case of a higher dimensional input, we flatten it to a vector. We first transform the input vector with an embedding layer and assign each input element/feature $x_j$ a unique vector via a local map $\phi(x_j)$. The feature map has a crucial role in determining the expressive power of one-layer tensor network models or, more generally, variational quantum machine learning models \cite{schuld2021effect}. We will use a linear embedding function of continuous features in the unit interval $x_j\in[0,1]$
\begin{align}
    \phi(j)=\begin{pmatrix}x_j\\1-x_j\end{pmatrix},
    \label{eq:embedding}
\end{align}
which will simplify the initialisation of the deep tensor network.

Then, we transform the embedded vector with a linear map. We represent the map in the form of a tensor network. Depending on the task, this tensor network is either a matrix product state (MPS) \cite{stoudenmire2016supervised,efthymiou2019tensornetwork}, or a matrix product operator (MPO) \cite{guo2018matrix, guo2020tensor}.

For the classification task we use a matrix product state
\begin{align}
    \psi^c_{s_1,s_2\ldots s_N} &= \sum_{a_0,\ldots,a_N=1}^{{D}_{\rm MPS}}{A}^{s_1}_{a_0,a_1}(1){A}^{s_2}_{a_1,a_2}(2)\ldots C^c_{a_{N/2},a_{N/2}}\ldots{A}^{s_N}_{a_{N-1},a_N}(N)\tilde{B}_{a_N,a_0},\\ s_j&=1,2,\ldots d,\quad \mbox{and}\quad c=1,2,\ldots \mbox{\#classes}.
\end{align}
The bond dimension $D_{\rm MPS}$ determines the sizes of the tensors $A(j)$,  $B$, and $C$ containing trainable parameters and is the only hyper-parameter of the MPS model. We calculate the probability for the class $c$ by contracting input embeddings $\phi(j)$ with the MPS $\psi^c$
\begin{align}
    p_c=\sigma\left(\sum_{s_1,s_2\ldots s_N=1}^d \phi_{s_1}(1)\phi_{s_2}(2)\ldots \phi_{s_N}(N) \psi^c_{s_1,s_2\ldots s_N}\right).
\end{align}
To obtain the final probabilities, we introduced a nonlinearity $\sigma$. There are several natural choices for $\sigma$. In the case of complex MPS tensors $A(j)$, we use the modulus squared $|\bullet|^2$ nonlinearity. In the case of real parameters, we use the standard softmax nonlinearity.

Sequence prediction requires the application of a matrix product operator \cite{guo2018matrix}
\begin{align}
    W_{s_1,s_2\ldots s_N}^{t_1,t_2,\ldots,t_N} = \sum_{a_0,\ldots,a_N=1}^{D_{\rm MPO}}M^{s_1,t_1}_{a_0,a_1}(1)M^{s_2,t_2}_{a_1,a_2}(2)\ldots M^{s_N,t_N}_{a_{N-1},a_N}(N)G_{a_N,a_0},\quad s_j,t_j=1,2,\ldots d.
    \label{eq:MPO full tensor}
\end{align}
The MPO bond dimension is denoted by $D_{\rm MPO}$. The core tensors $M(j)$ and the boundary matrix $G$ are parameters of the model and can be, as in the MPS case, real or complex. We obtain the predicted sequence by applying the linear operator $W$ to the tensor the embedding vectors $\phi(j)$
\begin{align}
    y_{t_1,t_2,\ldots t_N} = \sum_{s_1,s_2\ldots s_N=1}^d \phi_{s_1}(1)\phi_{s_2}(2)\ldots \phi_{s_N}(N)W^{t_1,t_2,\ldots t_N}_{s_1,s_2\ldots s_N}.
    \label{eq:mpo-mps}
\end{align}
In the last step we decode the obtained tensor $y_{t_1,t_2,\ldots t_N}$ to a sequence vector. In \cite{guo2018matrix} this is done in two steps: i) reducing the bond dimension of the output with an iterative procedure, and ii) decoding the $D_{\rm MPS}=1$ MPS embeddings into a sequence by applying an inverse of the embedding step (i.e. $\phi^{-1}$). 

The tensors in the described MPS and MPO layers can be position-independent, i.e. $A(j)=A$, $M(j)=M$. In this case, we say that the model is uniform. The uniform models have much fewer parameters. More importantly, we can apply the uMPS to inputs of different lengths. 

The main drawback of the described application of the matrix product operators is that they increase the bond dimension of the matrix product state. Therefore, they require a sweeping procedure, with the SVD as the essential computational tool, to reduce the bond dimension $D_{\rm MPS}$ to 1. Multiple applications of matrix product operators in this form are thus not practical. We introduce a related but different MPO layer that does not increase the MPS bond dimension. Many successive MPO layers result in a deep tensor network that is exponentially more expressive than the shallow, one-layer tensor network discussed in \cite{guo2018matrix}.

\section{Deep tensor networks}
We will now introduce deep tensor networks. Our approach is motivated by the time-dependent variational principle with matrix product states \cite{haegeman2016unifying}, which is used to simulate the dynamics of many-body quantum systems. Since our embedding is an MPS with $D_{\rm MPS}=1$, the procedure outlined in \cite{haegeman2016unifying} simplifies even further. First, we will introduce the MPO layer, the fundamental building block of deep tensor networks. Then, we will discuss how many MPO layers can be combined to form a deep tensor network. Finally, we will compare the MPO layer to the linear dot-attention mechanism.
\subsection{Matrix product operator layer}
\label{sec:mps-tdvp}
\begin{figure}[!htb]
    \centering
    \includegraphics[width=0.75\textwidth]{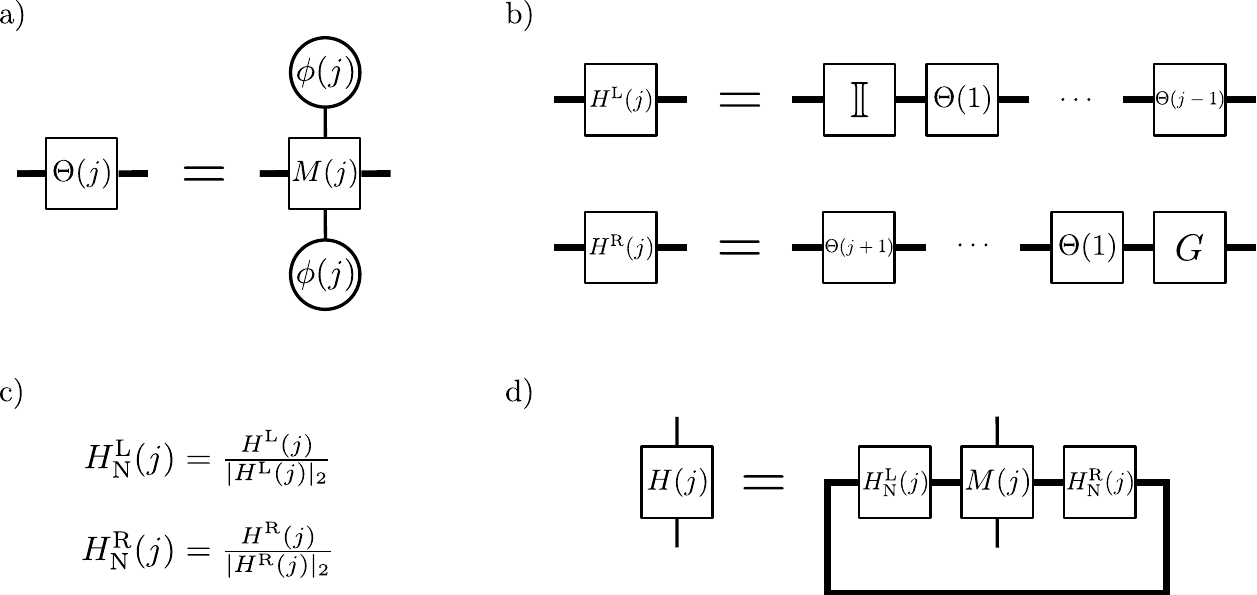}
    \caption{Diagramatic representation of the local weight matrix $H(j)$. {\bf a)} Calculation of the matrices $\Theta(j)$, \eref{eq:theta}. {\bf b)} Contraction of the left and the right context matrices \eref{eq:left context}  and \eref{eq:right context}. {\bf c)} Normalisation of the context matrices. {\bf d)} The final contraction in calculation of the local weight matrix \eref{eq:hmpo contraction}.}
    \label{fig:hmpo}
\end{figure}

The main idea is to transform the embeddings $\phi(j)$ as
\begin{align}
    \phi(j) \xrightarrow{\rm MPO} \psi(j)=\sigma\left(H(j)\phi(j)\right),
    \label{eq:tdvp update}
\end{align}
where $\sigma$ denotes a nonlinearity, and $H(j)$ a local linear maps or local weight matrices. To calculate the local weight matrices $H(j)$, we first contract the MPO kernels with two copies of the embeddings
\begin{align}
    \Theta_{a_{j-1},{a_{j}}}(j)=\sum_{s_j,t_j=1}^{d}\bar{\phi}(j)_{s_{j}}M^{s_{j},t_{j}}_{a_{j-1},a_j}(j)\phi(j)_{t_{j}},
    \label{eq:theta}
\end{align}
With bar $\bar{\bullet}$ we denote complex conjugation. Next, we use the contracted matrices $\Theta(j)$ to sequentially calculate the left and the right context matrices. The left context matrices are 
\begin{align}
    H^{{\rm L}}(1) &= \mathds{1}_{D_{\rm MPO}},\\
    H^{{\rm L}}(j)&=H^{{\rm L}}(j-1)\Theta(j-1)
    \label{eq:left context}
\end{align}
and the right context matrices are 
\begin{align}
    H^{{\rm R}}(N) &= G,\\
    H^{{\rm R}}(j)&=\Theta(j+1)H^{{\rm R}}(j+1).    
    \label{eq:right context}
\end{align}
The boundary conditions $H^{\rm L}(1)$ and $H^{\rm R}(N)$ are chosen such that their product gives the boundary matrix $H^{\rm R}(N) H^{\rm L}(1) = G$, which is defined in \eref{eq:MPO full tensor} and determines the boundary conditions of the matrix product operator. To avoid overflow/underflow for long inputs we normalise the left and right context matrices $H^{\rm L,R}_{\rm N}(j)=H^{\rm L,R}(j)/|H^{\rm L,R}(j)|_2$. Finally, we calculate the local weight matrix by contracting the normalised left and right context matrices with the corresponding MPO tensor
\begin{align}
    H^{s,t}(j) = {\rm Tr}\left(H^{\rm L}_{\rm N}(j)M^{s,t}(j)H^{\rm R}_{\rm N}(j)\right)
    \label{eq:hmpo contraction}
\end{align}
A diagrammatic representation of the $H(j)$ calculation is shown in \fref{fig:hmpo}.

The outlined procedure differs from the original MPS-TDVP \cite{haegeman2016unifying}. We do not restrict the embeddings to be $L_2$ normalised, but rather normalise the left and right context matrices. The normalisation enables us to work with a more general MPO and makes the algorithm stable for longer sequences. Since we are working with $D_{\rm MPS}=1$ matrix product states, we can void one step in the original algorithm. Also, we do not apply the local transformations \eref{eq:tdvp update} sequentially but in parallel. Finally, we are free to apply any nonlinearity. In the original method, the matrix exponential of the local weight matrices (effective Hamiltonians) without a nonlinearity is used to update the MPS kernels. We will apply the \textit{matrix exponential} $\exp H(j)$ with a linear activation or the $H(j)$ with a \textit{sigmoid} nonlinearity.

The MPO layer can be applied efficiently. The computationally expensive step is the calculation of the final contractions \eref{eq:hmpo contraction} which scales as $\mathcal{O}\left(N d^2 D{_{\rm MPO}^2}{\rm rank}\,G\right)$. 

\subsection{Multiple MPO layers}
\label{sec:deeptn}
\begin{figure}[!htb]
    \centering
    \includegraphics[width=0.3\textwidth]{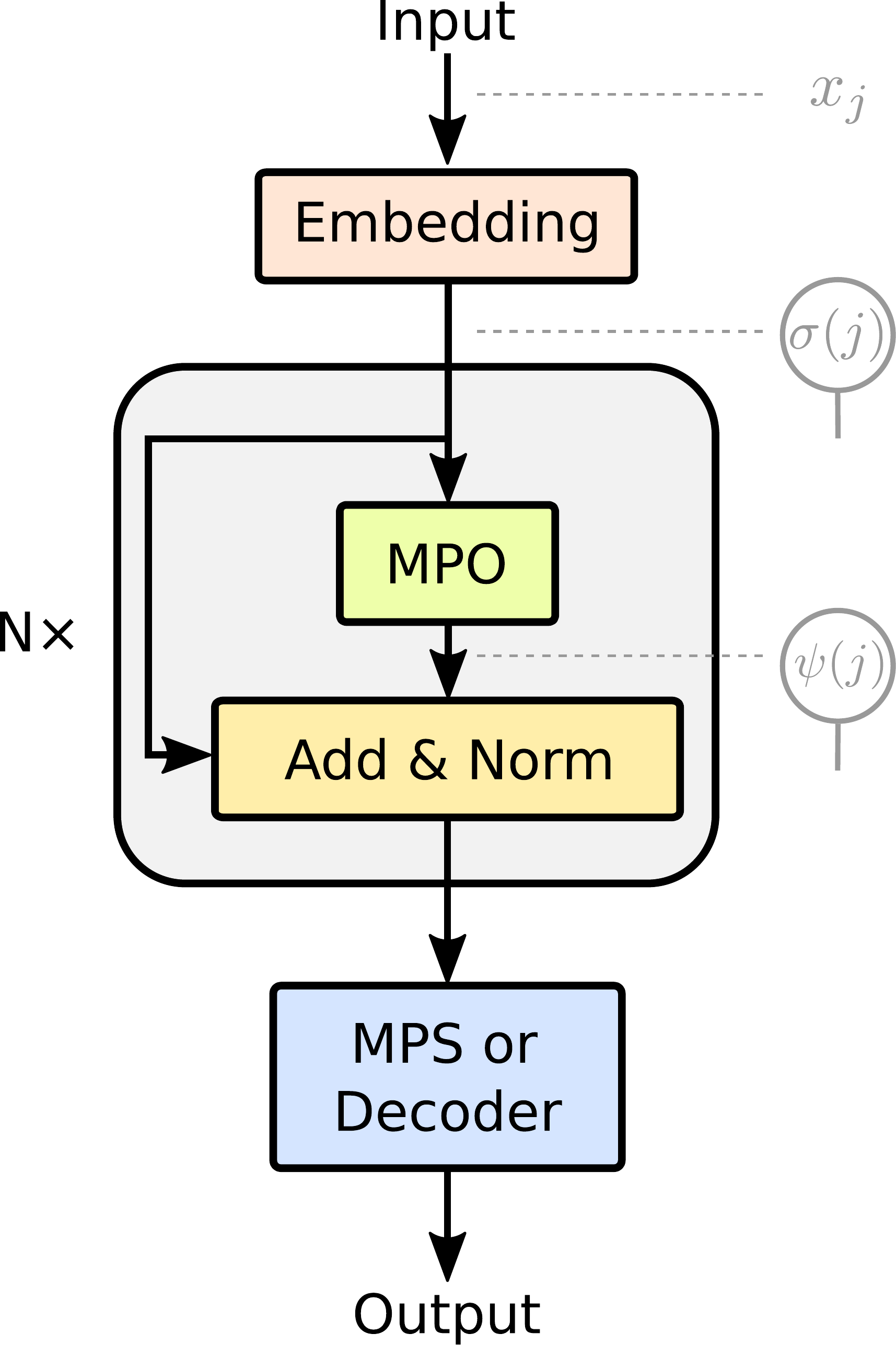}
    \caption{Schematic representation of a deep tensor network. }
    \label{fig:dtn_full}
\end{figure}

The main benefit of the proposed method is that the bond dimension of the input matrix product state (the embedding vectors) does not increase with an application of the MPO layer. Therefore, we can apply the MPO layer repeatedly. We call the resulting model a deep tensor network. The final architecture is shown in \fref{fig:dtn_full} and consists of an embedding layer followed by the MPO layers. We also add a skip connection and normalisation. Both can be turned off and add two additional hyperparameters. The final layer depends on the task. In the classification case, we use the MPS layer described in \sref{sec:tnml}, and in the sequence prediction case, the final layer is a decoder layer. The decoder is a simple inverse operation of the embedding layer.

\subsection{Relation to standard feed forward networks}
\label{sec:tn-nn}
In this section, we will compare the standard feed-forward neural networks with the introduced MPO layer. In particular, we will focus on the attention mechanism proposed in \cite{vaswani2017attention}. 

The attention layer transforms embedding vectors $\phi(j)$ as follows. First, embeddings are linearly transformed into a query, key, and value triplets $\phi(j)\rightarrow(q_j,k_j,v_j)$. The attention output is then calculated as
\begin{align}
    \psi(j)={\rm softmax}\,\left(\frac{q_jK}{\sqrt{d}}\right)V^T,
    \label{eq:attention}
\end{align}
where $K=(k_1,k_2,\ldots k_N)$ and $v=(v_1,v_2,\ldots v_N)$. If we remove the softmax nonlinearity from \eref{eq:attention} we can express the resulting transformation as an application of the MPO layer with a bond dimension $D_{\rm MPO} = d^2+1$ (for  details see \aref{app:permutation mpo}). Therefore, we can interpret the proposed MPO layer as generalised linear dot attention. It can model higher-order correlations/interactions between embeddings and explicitly incorporates the embedding position. On the other hand, we can not include a nonlinearity of type \eref{eq:attention} into the formalism. 

\section{Image classification with deep tensor networks}
\label{sec:MPS classification}
We will now discuss the application of deep tensor networks on the MNIST and Fashion MNIST image classification tasks. Tensor networks have previously been applied to both tasks with moderate success \cite{stoudenmire2016supervised,stoudenmire2018learning, efthymiou2019tensornetwork, liu2019machine,martyn2020entanglement,meng2020residual,chen2021residual,kong2021quantum, cheng2021supervised}. The best results using tensor network methods in combination with neural networks are 0.69\% error rate on MNIST \cite{cheng2021supervised} (CNN+PEPS) and 8.9\% error rate on Fashion MNIST \cite{cheng2021supervised} (CNN+PEPS). However, previous approaches did not use image augmentation and hyperparameter search. To obtain our matrix product state baseline models we first perform extensive hyperparameter search. We study the performance of the models on the normal and permuted datasets. Finally, we compare the obtained results with the results for deep tensor networks.

\subsection{Baseline MPS models}
\label{sec:MPS baselines}
We perform image classification with the MPS models in two steps as discussed in \sref{sec:tnml}. We chose the initial condition such that the logits are close to one for any input $x_j$. In particular we choose $[A^{s_j}(j)]_{ij}=\mathds{I}_{D_{\rm MPS}}+R_j$, where $\mathds{I}_{\rm MPS}$ denotes a $D\times D$ identity matrix and $R_j$ a $D\times D$ matrix with normally distributed elements with mean zero and variance $\epsilon$. During training, we use the standard cross-entropy loss with L2-regularization of the MPS matrices. In our final setting, we use batch sizes from 128 to 512 (depending on the memory consumption), AdamW optimizer, and reduce the learning rate on a plateau with $\gamma=0.5$ and patience 20 or 30 epochs. We determine the optimizer as a part of the hyperparameter search.  

We use \textit{torchvision} to augment the dataset images. We performed hyperparameter search for the various transformation parameters and probabilities. In particular, we use the following sequence of transformations: random pixel shift, random colour jitter, random sharpness, random Gaussian blur, random horizontal flip, random affine, random perspective, resize, crop, random elastic transformation, random erasing. We also tune the probability of applying each random transformation while keeping the order of transformations fixed. During hyperparameter search, we restrict the training to a maximum of 30 epochs. For more details on hyperparameter search consult \aref{app:hyperparameters}.

\subsection{Baseline results}
After finding the best hyperparameters, we train an MPS model and a uniform MPS (uMPS) model with increasing bond dimensions. We evaluate the models on the MNIST and Fashion MNIST datasets. For each dataset, we consider two cases: 1) classification of images and 2) classification of permuted images. In the second case, we flatten the image to a vector and then apply a fixed random permutation. Additionally, we evaluate the performance of an ensemble obtained by averaging logits of the 10-fold cross-validation models. On the permuted datasets, we consider ensembles obtained from models with the same perturbation of the inputs (\textit{ensemble I}) and ensembles obtained from models with different perturbations of the inputs (\textit{ensemble II}).

\begin{figure}[!htb]
    \centering
    \includegraphics[width=0.23\textwidth]{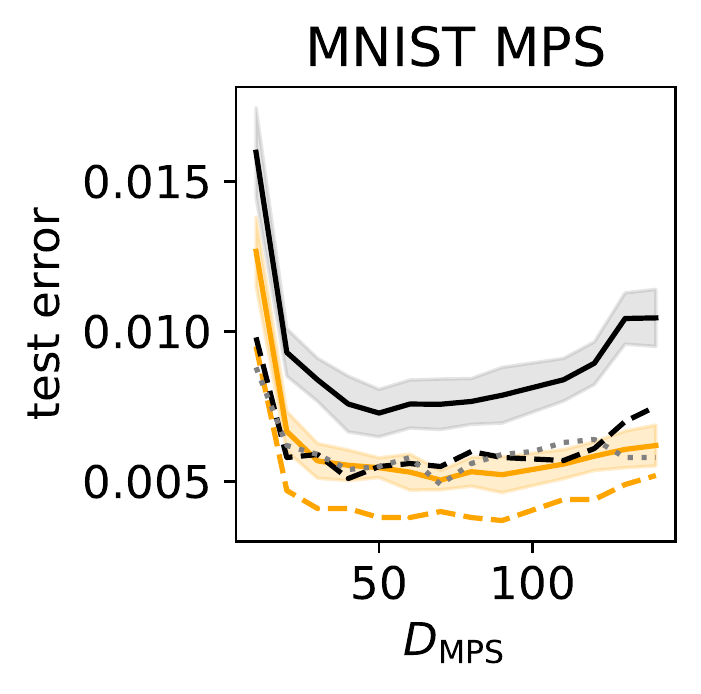}
    \includegraphics[width=0.23
    \textwidth]{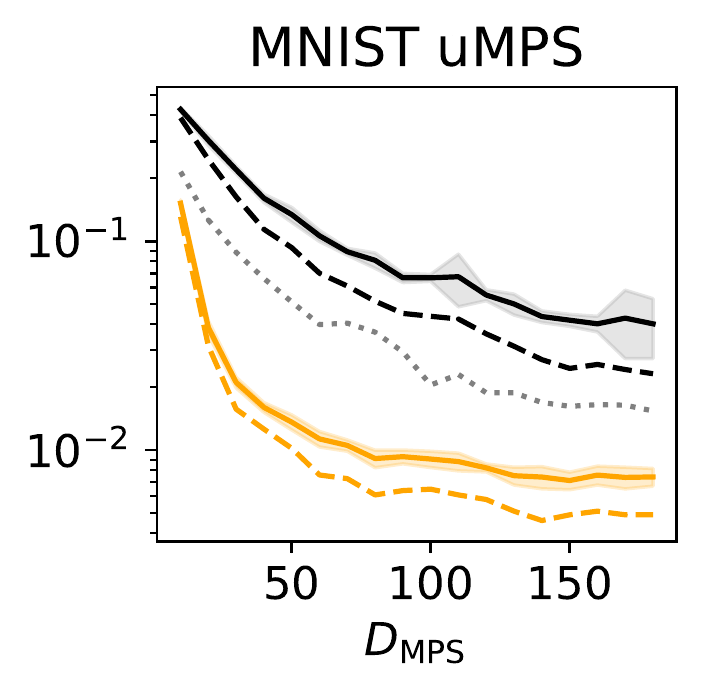}
    \includegraphics[width=0.23\textwidth]{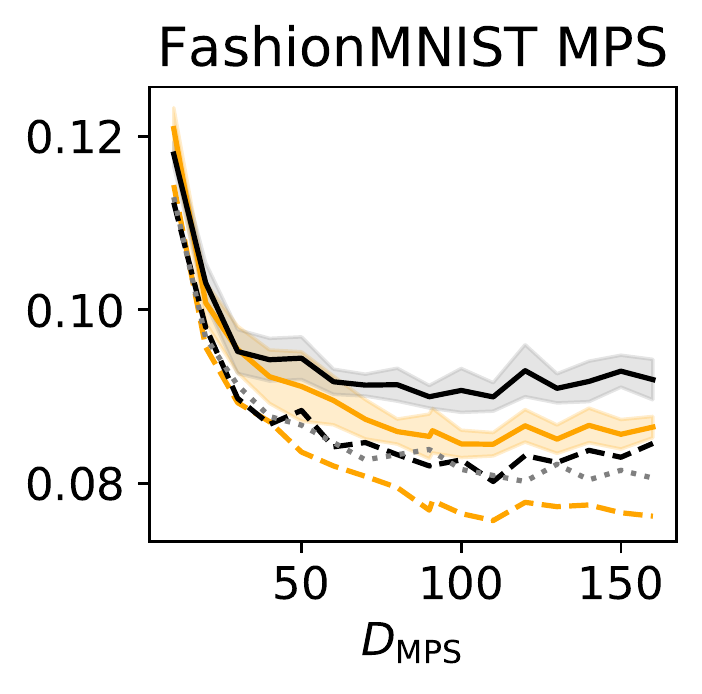}
    \includegraphics[width=0.253
    \textwidth]{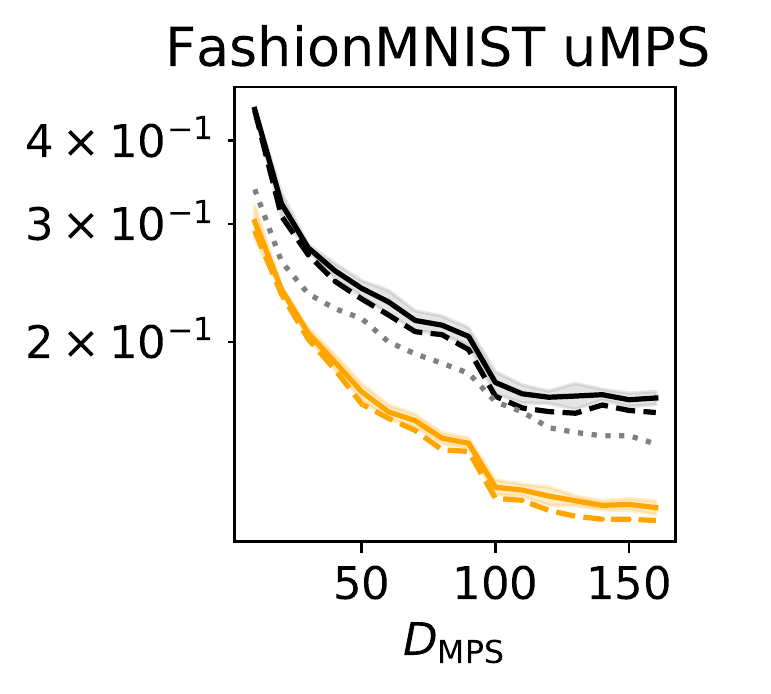}
    
    \includegraphics[width=\textwidth]{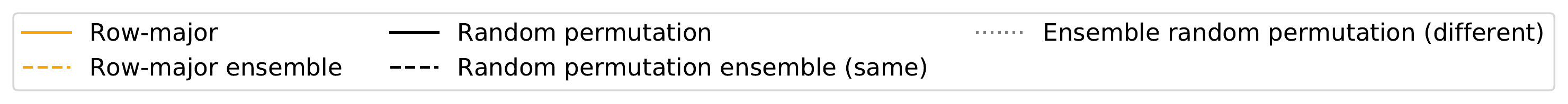}
    
    \caption{Test error of MPS and uMPS models. The orange lines represent the results on standard datasets, and the black lines correspond to permuted datasets. The grey and orange shaded regions denote the standard deviation of the 10-fold cross-validation.}
    \label{fig:image accuracy}
\end{figure}

We evaluate all baseline MPS models with increasing bond dimensions, see \fref{fig:image accuracy}. The MPS models achieve the optimum performance already for relatively small bond dimensions $D\approx70$ (MNIST) and $D\approx100$ (Fashion MNIST). If we increase the bond dimension further, the performance even decreases in some cases. On the other hand, the results improve if we use ensemble predictions. Since an MPS model is a linear model in the exponentially large feature space, an ensemble of MPS models is again an MPS model with a larger bond dimension ($D_{\rm emsemble} = D_{\rm model~1} + D_{\rm model~2}+\ldots D_{\rm model~10}$). The discrepancy between accuracy saturation with increasing bond dimension $D_{\rm MPS}$ and better ensemble results suggests that the increase of the test error with increasing bond dimension is a consequence of our training procedure. One possibility to improve the training is to train an ensemble of MPS models with a small bond dimension and then combine it to a larger model, which would only be fine-tuned (i.e. trained with a much smaller learning rate). We can combine the ensembling procedure with the SVD compression to reduce the bond dimension of the final model. It would be interesting to see if the sequential application of the ensembling and compression improves the accuracy of final baseline models.

The performance of the MPS models on the permuted datasets qualitatively follows the performance on the standard datasets. However, we observe an overall increase in the test error. Nevertheless, the ensemble test error on the permuted datasets matches or is even smaller compared to the test error of single models on the non-permuted/standard datasets. Finally, we do not observe a significant difference between the ensembles I and II on the permuted datasets.  

In the case of uniform MPS models, the test error saturates at larger bond dimensions $D\approx 170$ and is larger as in the MPS case. We also observe a larger difference in the test error between the permuted and non-permuted datasets. In contrast to the MPS case, we observe in the uMPS models a difference in the studied ensembling procedures I and II. The ensemble of models with different permutations (\textit{ensemble I}) has a smaller error than the ensemble of models with the same permutation (\textit{ensemble II}). This difference might be due to an in-equivalence of permutations for an image. For example, one permutation might be better for classifying the class \textit{A} and another permutation might be better to classify the class \textit{B}. Having different permutations in an ensemble could therefore lead to significant improvements. However, it is unclear if our interpretation is correct and why there is no difference in the MPS models. We leave these questions for future research since our primary goal is to establish good MPS baselines. 

\begin{figure}[!htb]
    \centering
    \includegraphics[width=0.4\columnwidth]{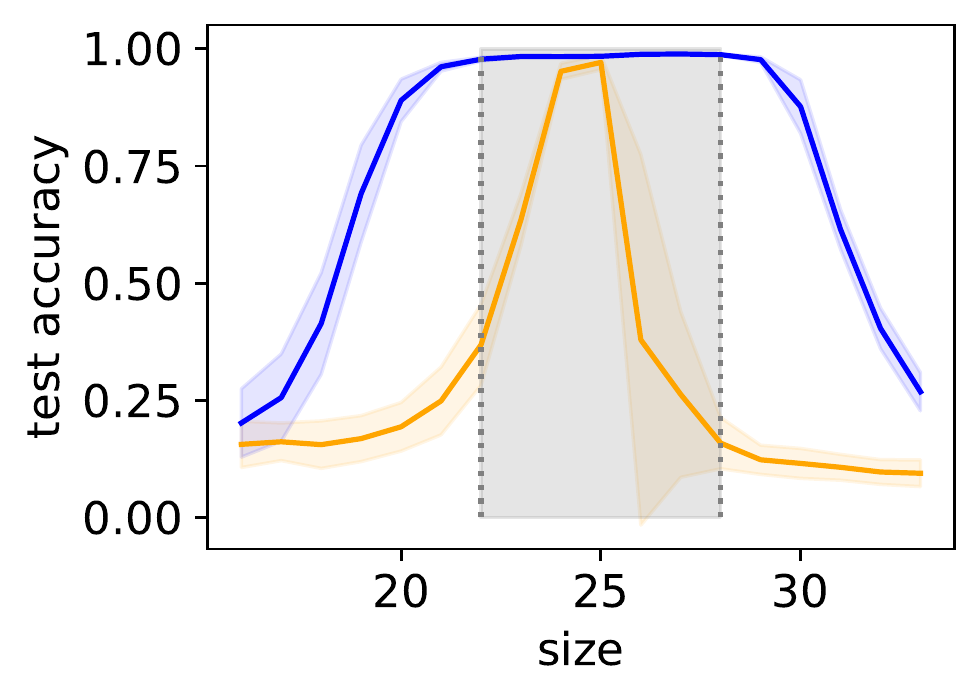}
    \includegraphics[width=0.4\columnwidth]{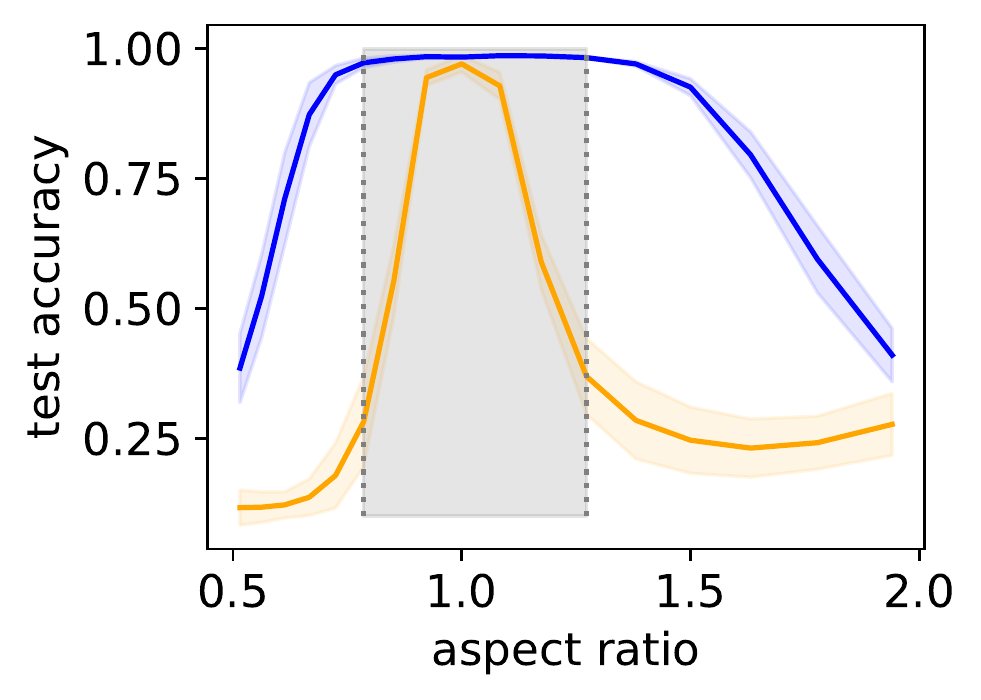}
    
    \includegraphics[width=0.55\textwidth]{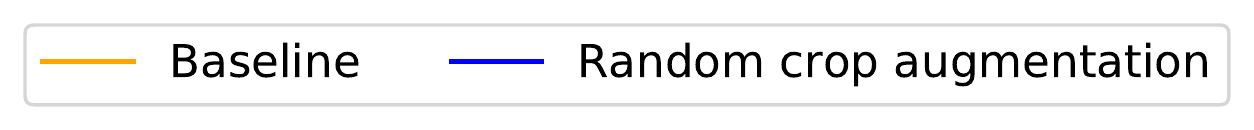}
    \caption{Robustness of uMPS models to size and aspect ratio changes. The orange line represents results for training with only one size, and the blue line corresponds to training with random image width and height in the range from 23 to 28. The gray shaded regions show the range of training sizes (left) and aspect ratios (right).}
    \label{fig:image size invariance}
\end{figure}

In addition to the accuracy, we also study the robustness of the uMPS to image size and aspect ratio changes. We evaluate two training procedures. First, we train the models only with the image size determined by the hyperparameter search. Second, we vary also the input size and aspect ratio. As shown in \fref{fig:image size invariance} the accuracy of models trained with only one size drops if we change the input size or aspect ratio. In contrast, models trained on different input sizes are robust against the varying input sizes even outside the range provided during training (up to $\pm12.\%$ of the original input). However, if we further change the input sizes or aspect ratios, we again observe a drop in accuracy.

We present a summary of our best baseline MPS models together with the deep tensor network results (discussed in the next section) in table \ref{tab:classification results}. Our baseline ensemble models achieve comparable results to the neural network and mixed (tensor + neural network) approaches on the MNIST and FashionMNIST datasets \cite{glasser2018supervised,meshkini2019analysis}. Besides, our baseline single model results indicate that linear mutual information scaling with the system size might not be an excluding indicator for good classification performance, as suggested in \cite{lu2021tensor}. Namely, the accuracy of MPS models does not drop dramatically on the permuted datasets, where we expect a volume-law scaling.

\subsection{Deep tensor network results}
We will now study the impact of one or more MPO layers on the performance of the baseline models discussed in the previous section. The MPO layer introduced in \sref{sec:deeptn} has several additional hyperparameters. Along with the bond dimension $D_{\rm MPO}$ we can tune the number of layers, the non-linearity, normalisation of the output, and residual connections. We tried several configurations but did not perform an additional hyperparameter search. Therefore, it is likely that the deep tensor network results presented in the following can be further improved. 

\begin{figure}[!htb]
    \centering
    \includegraphics[width=0.45\textwidth]{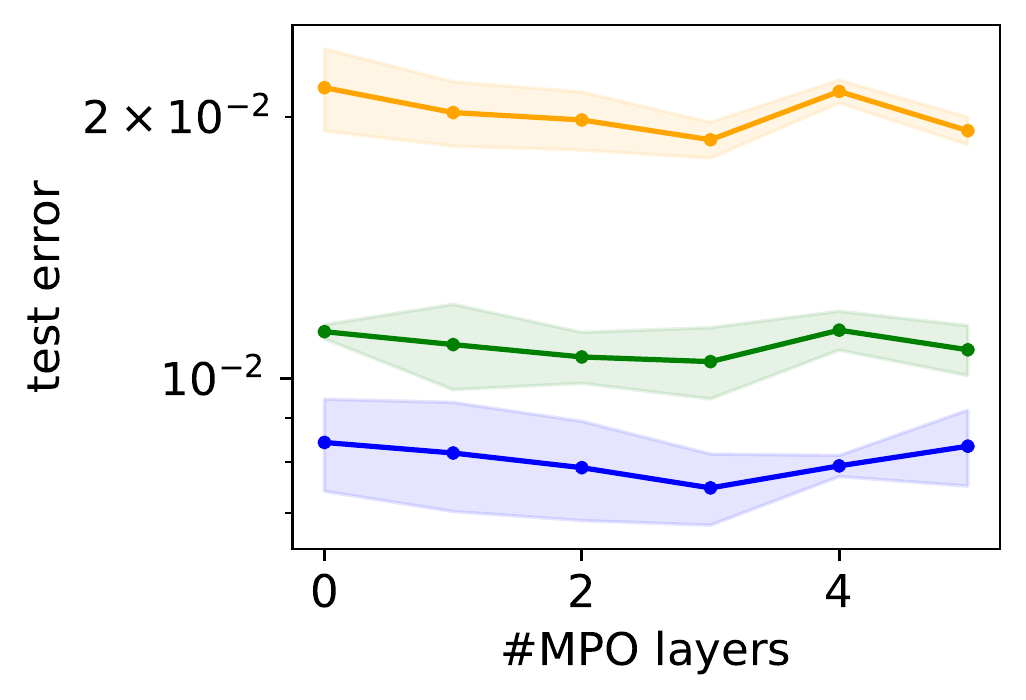}
    
    \includegraphics[width=0.50\textwidth]{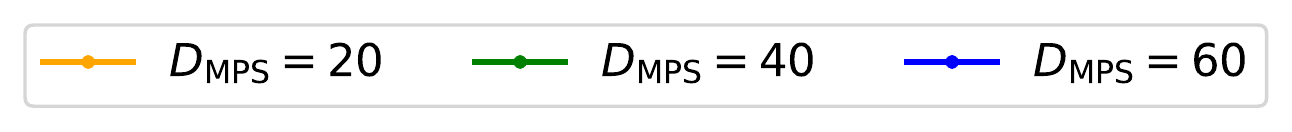}
    \caption{MNIST test accuracy of deep tensor networks as a function of the depth. Independent of the size of the initial model, additional uMPO layers of fixed size $D_{\rm uMPO}=10$ improve the accuracy. If the model becomes too deep, training becomes more difficult, and the error increases. Shaded regions denote the standard deviation of the 10-fold cross-validation.}
    \label{fig:tdvp depth}
\end{figure}

We first studied the effect of the network depth on the model accuracy. We focused on the MNIST dataset and uMPO layers without residual connection, linear activations, without the output normalisation, and the bond dimension $D_{\rm uMPO}=10$. In \fref{fig:tdvp depth} we show that the accuracy increases, although only slightly, with the number of layers up to depth 3, then it decreases by adding more layers. The main reason for this performance drop with increasing depth is that we do not fine-tune training parameters, which is critical for deeper models.

\begin{figure}[!htb]
    \centering
    \includegraphics[width=0.4\textwidth]{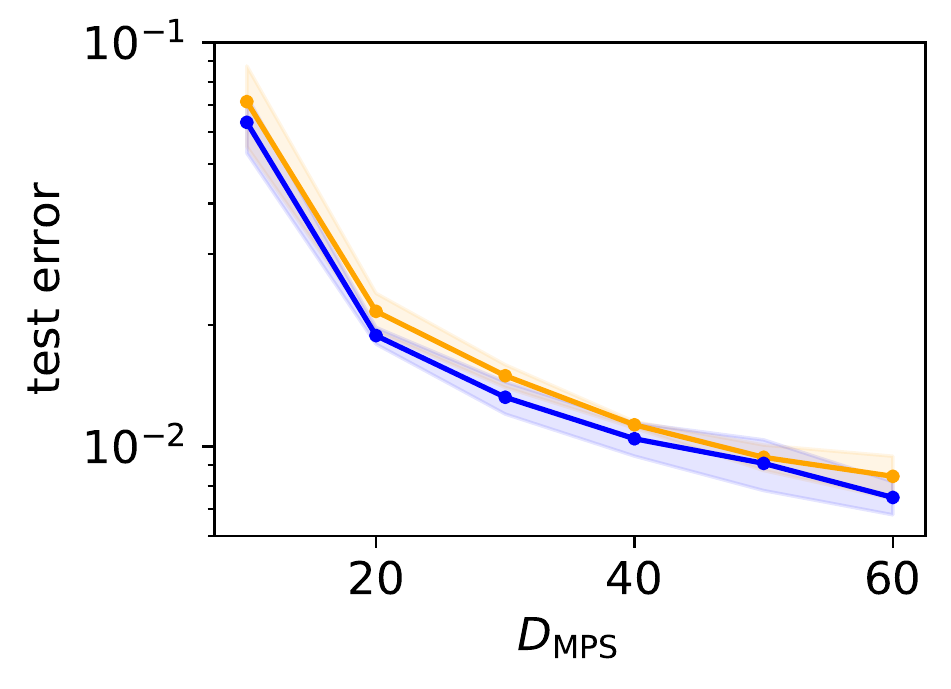}
    
    \includegraphics[width=0.4\textwidth]{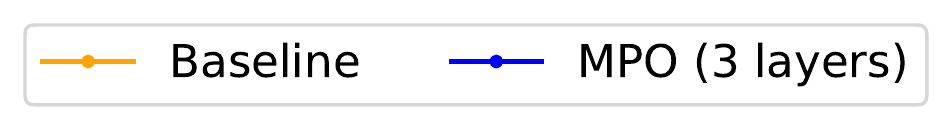}
    \caption{MNIST test accuracy of deep tensor networks as a function of the baseline model size. Shaded regions denote the standard deviation of the 10-fold cross-validation.}
    \label{fig:tdvp D}
\end{figure}

Next, we focus on the model with three uMPO layers and fix the bond dimension of the MPO layer to $D_{\rm uMPO}=10$. We then increase the bond dimension of the baseline model $D_{\rm MPS}$ from 10 to 60. We observe that the deep tensor network model decreases the error rate in all cases (see \fref{fig:tdvp D}). For $D_{\rm MPS}=60$ the number of parameters of the deep model is increased by around $8\%$ while decreasing the error by around $12\%$. 

\begin{figure}[!htb]
    \centering
    \includegraphics[width=0.4\textwidth]{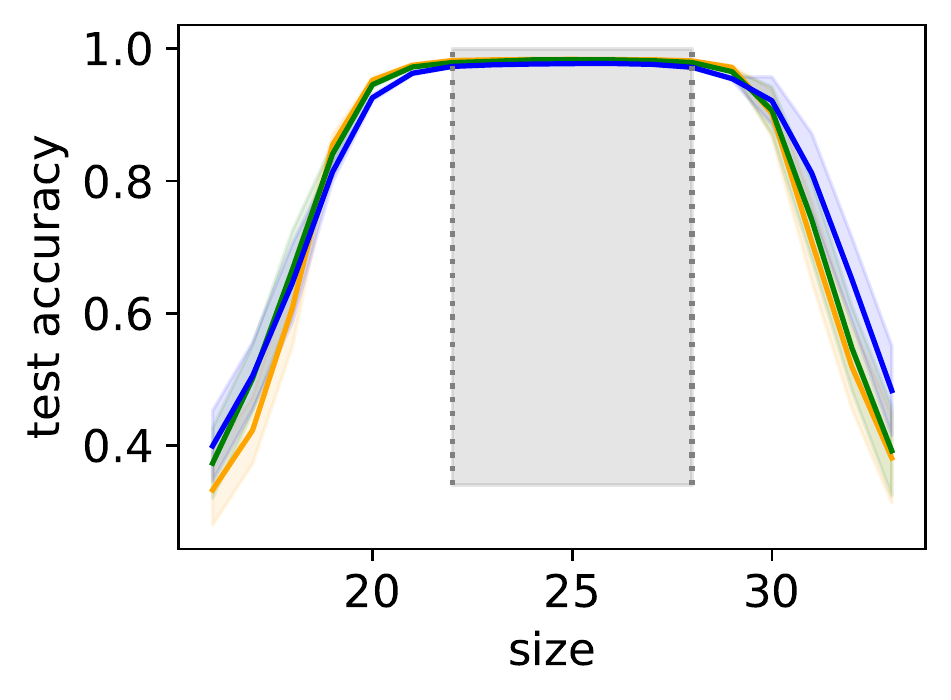}
    \includegraphics[width=0.4\textwidth]{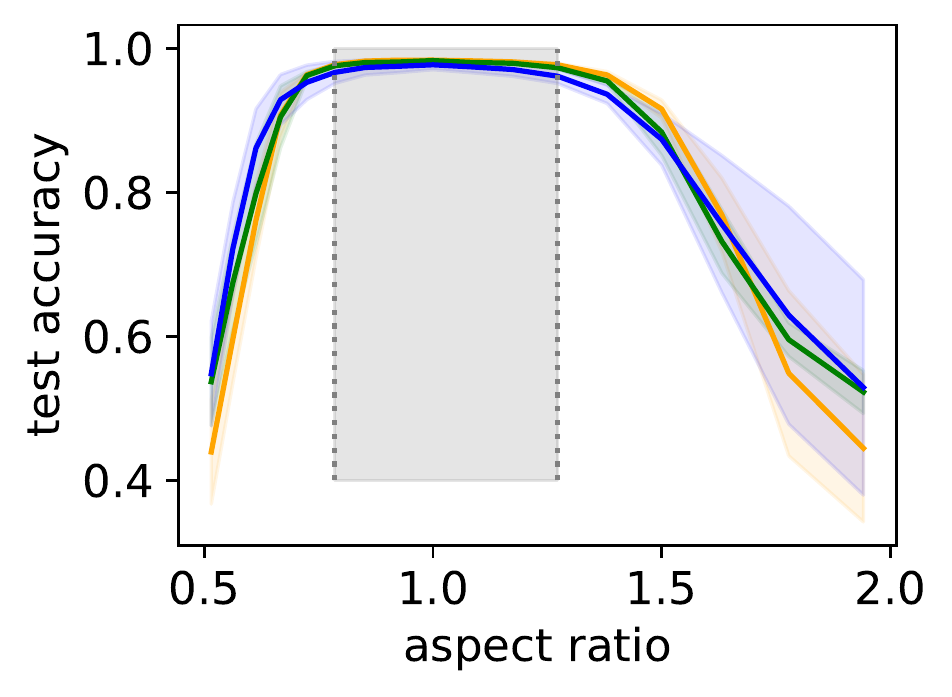}
    
    \includegraphics[width=0.7\textwidth]{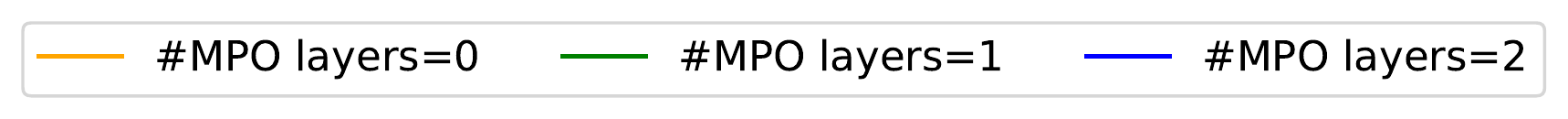}
    \caption{Robustness of uniform deep tensor networks to input aspect ratio and size changes on the MNIST dataset. The grey shaded regions denote the training-time range of sizes and aspect ratios. Coloured shaded regions denote the standard deviation of the 10-fold cross-validation.}
    \label{fig:duMPS robustness}
\end{figure}

We also evaluate the robustness of the deep uMPS models to input aspect ratio and size changes. Similarly, as in the MPS models, we observe that the models trained with a fixed size and aspect ratio are not robust to these changes. However, by varying the input size and aspect ratio during training the models become robust also to changes that are larger compared to the changes during training. In \fref{fig:duMPS robustness} we show the accuracy of models with zero, one, and two MPO layers. Interestingly, models with more layers are more robust to aspect ratio and size changes outside the training range.

Finally, we report the best results obtained by deep tensor network models with one MPO layer. We experimented with the linear, relu, sigmoid, and matrix exponential activation functions and found that the last case performs best. The drawback of calculating the matrix exponential is that it is slow. Therefore, we used an exact formula for $d=2$ and significantly decreased the computation time. We also tested models with and without the residual connections and different output normalisations. We find that the residual connections increase training speed whereas the output normalization does not have a significant effect on the model performance. Besides mentioned tests, we did not perform another hyperparameter search, which could further improve our results. 

Our single model deep tensor networks improve our MPS baselines in all tasks. We observe the largest decrease of the test error on permuted datasets with uniform models. This is expected since additional layers significantly widen the receptive field of the final layer, which is crucial in the permuted case due to long-range correlations. In the permuted case, we use only ensembles with different permutations since the tests on the MPS models revealed that this significantly improves the results. We summarise the final results in the table \ref{tab:classification results}.

\begin{table}[!htb]
    \centering
    \begin{tabular}{|l | c | c | c | c |} 
     \hline
      & MNIST (\%) & pMNIST (\%)& FashionMNIST (\%)& pFashionMNIST (\%) \\
        \hline
        Previous best TN \cite{cheng2021supervised}& 0.69  & /  & 8.9  & / \\
        \hline
        \hline    MPS  & 0.50 $\pm$ 0.03 & 0.73 $\pm$ 0.08 & 8.45 $\pm$ 0.13 & 9.00 $\pm$ 0.13 \\ 
        \hline
        MPS Ens. Same & {\bf 0.37} & {\bf 0.46} & 7.57 & 8.02 \\ 
        \hline
        MPS Ens. Rand & / & 0.49 & / & 8.27 \\ 
        \hline
        DTN & {\bf 0.49 $\pm$ 0.05} & {\bf 0.66 $\pm$ 0.05} & {\bf 8.28 $\pm$ 0.14} & { \bf 8.85 $\pm$ 0.14 } \\
        \hline
        DTN Ens. & 0.39  & 0.57 & {\bf 7.28} & {\bf 8.0} \\
        \hline
        \hline
        uMPS  & 0.71 $\pm$ 0.07 & 4.02 $\pm$ 1.28 & 11.23 $\pm$ 0.26 &  16.37 $\pm$ 0.38 \\ 
        \hline    
        uMPS Ens. Same & 0.46 & 2.32 & 10.80 & 15.63 \\ 
        \hline    
        uMPS Ens. Rand & / & 1.69 & / &  14.04 \\ 
        \hline  
        uDTN & {\bf 0.67 $\pm$ 0.06} & {\bf 3.64 $\pm$ 0.54} & {\bf 10.57 $\pm$ 0.39} & {\bf 15.06 $\pm$  0.75} \\
        \hline
        uDTN Ens. & {\bf 0.39}  & {\bf 1.66} & {\bf 9.84} & {\bf 13.61} \\
     \hline
    \end{tabular}
    \caption{Test error for models trained with the best hyperparameters. Besides the average error, we also report the standard deviation of the 10-fold cross-validation. We used one MPO layer for DTN models. In the non-uniform case we used $D_{\rm MPS}=60$ and $D_{\rm MPO}=10$ and in the uniform case we used $D_{\rm MPS}=150$ and $D_{\rm MPO}=10$. The ensembles of DTN in the permuted case were constructed by using different permutations of the inputs for each model.}
    \label{tab:classification results}
\end{table}

\section{Sequence prediction with deep tensor networks}
\label{sec:sequence}
In this section, we evaluate the proposed deep tensor networks on the sequence prediction task. We will consider an algorithmic sequence to sequence transformation determined by cellular automata. Cellular automata are a universal discrete space-time dynamical system with a finite set of possible states at each position \cite{wolfram1983statistical, wolfram2002new}. We define a cellular automata (CA) by a set of rules which transform one configuration of states into another configuration. We will consider the rule 30 one-dimensional automata, which exhibits chaotic behaviour. We can express the rule 30 automata as an MPO transformation with a bond dimension $D_{\rm MPO}=4$ \cite{guo2018matrix}. To gradually increase the problem difficulty, our task will be to predict the state after $j$ applications of rule 30. Due to chaoticity, we expect that the bond dimension of a single MPO describing the $j$-step transition grows exponentially with $j$, i.e. $D_{\rm MPO}(j) \approx 4^j$. On the other hand, a deep tensor network model can solve this problem by repeating one layer that describes a single step of the rule $j$ times. Therefore, deep tensor networks are exponentially more efficient than shallow one-layer tensor networks.

The local rule 30 is uniform. Hence, we will consider only uMPO models. In contrast, we can not use the approach proposed in \cite{guo2018matrix} to construct uMPO models due to a sweeping optimisation procedure that relies on the singular value decomposition. We utilise the model depicted in \fref{fig:hmpo} with a linear embedding layer defined in \eref{eq:embedding}. The embedding layer is followed by one or more MPO layers. The final layer is an encoding layer, which returns the first component of the L1-normalised output of the last MPO layer. During training, we minimise the L2 distance between the actual sequence and the predicted sequence. During the evaluation, we round the predicted number to the closest integer and calculate the accuracy as the percentage of correctly predicted states. We say that we solve the problem if the predicted sequences match the actual sequences for all considered test cases.

Similarly to the image classification case, we experimented with the activation functions, residual connections, and MPO output normalisation. We found, that we can solve the problem of predicting the next sequence for $j=1$ for a finite fixed input size (e.g. $N=30$) by using \textit{sigmoid} activation functions and the bond dimension $D_{\rm MPO}=2$. Our trained solution for the standard rule 30 automata is more compact than the exact solution presented in \cite{guo2018matrix}. We can understand this by adopting the probabilistic interpretation. The output of our models are probabilities for the states to be 0 or 1. To have the correct prediction, we only require that the probability for the correct value is larger than the probability for the wrong value. Nevertheless, it is interesting to observe that already for this simple task we find a more compact solution as the analytic MPO representation of the rule 30 \cite{guo2018matrix}. In contrast, we were unable to solve the $j=1$ problem with \textit{matrix exponential} activation and even $D_{\rm MPO}=4$. However, by adding one more layer with \textit{matrix exponential} activation we solved the $j=1$ problem with $D_{MPO}=2$. 

\begin{figure}[!htb]
    \centering
    \includegraphics[width=0.245\textwidth]{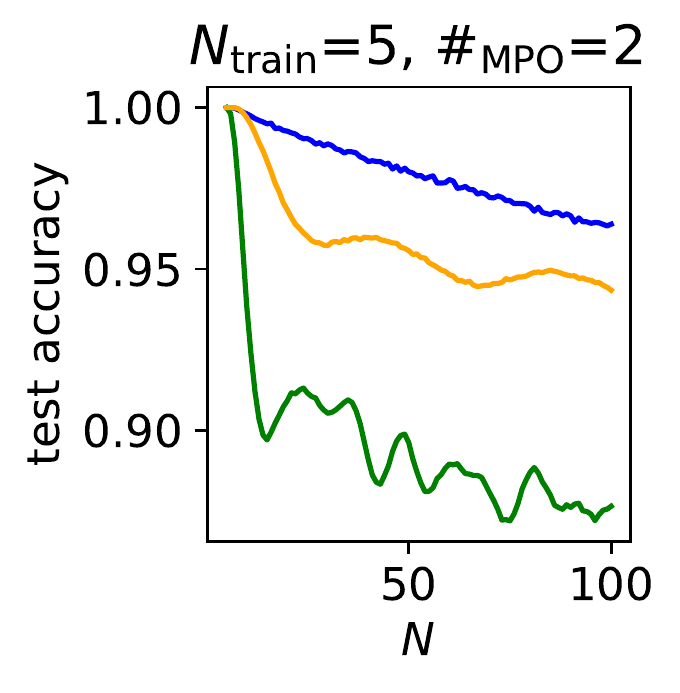}
    \includegraphics[width=0.245\textwidth]{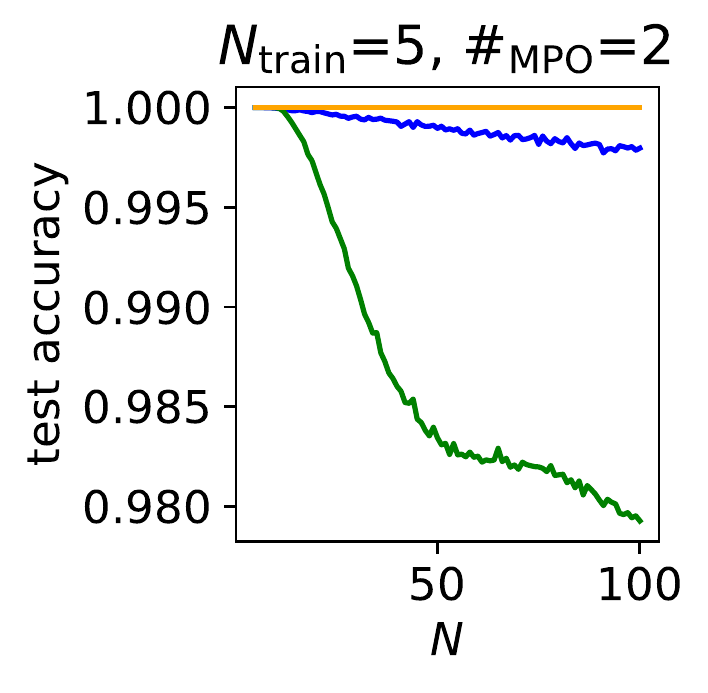}
    \includegraphics[width=0.245\textwidth]{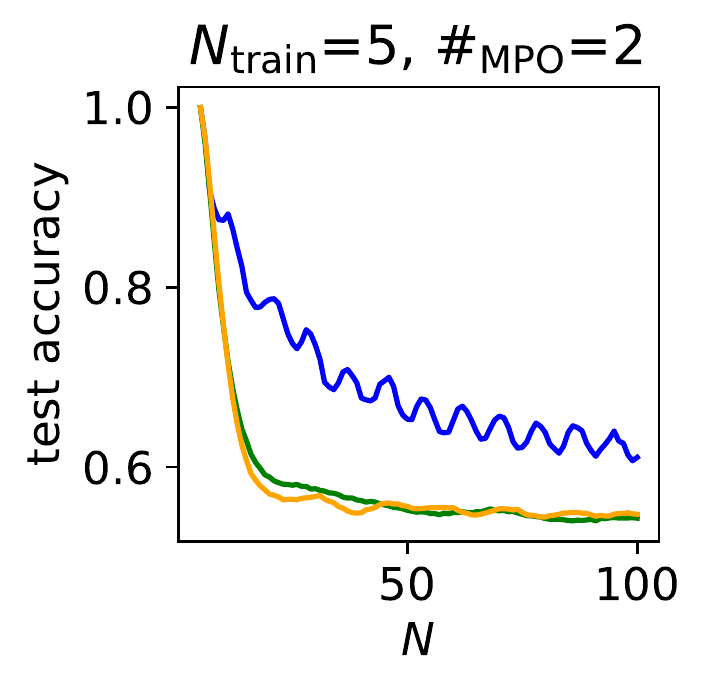}
    \includegraphics[width=0.245\textwidth]{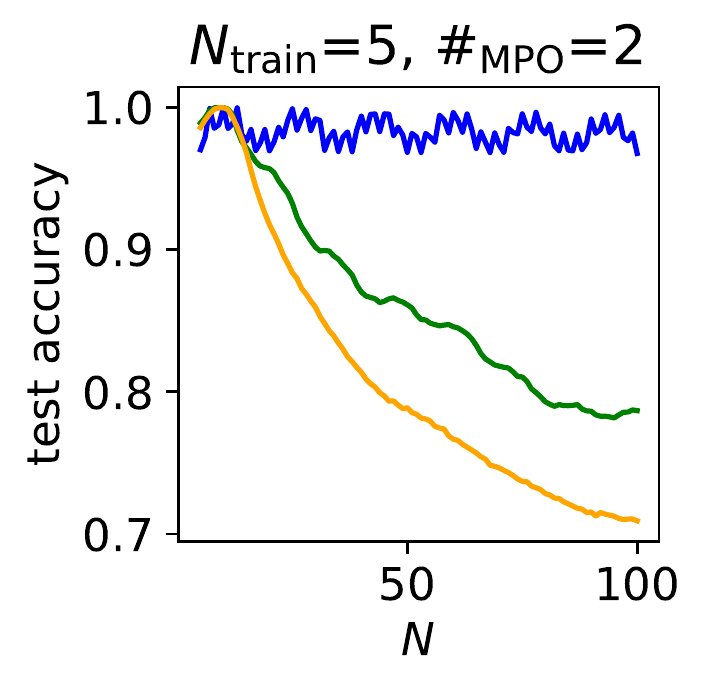}

    \includegraphics[width=0.245\textwidth]{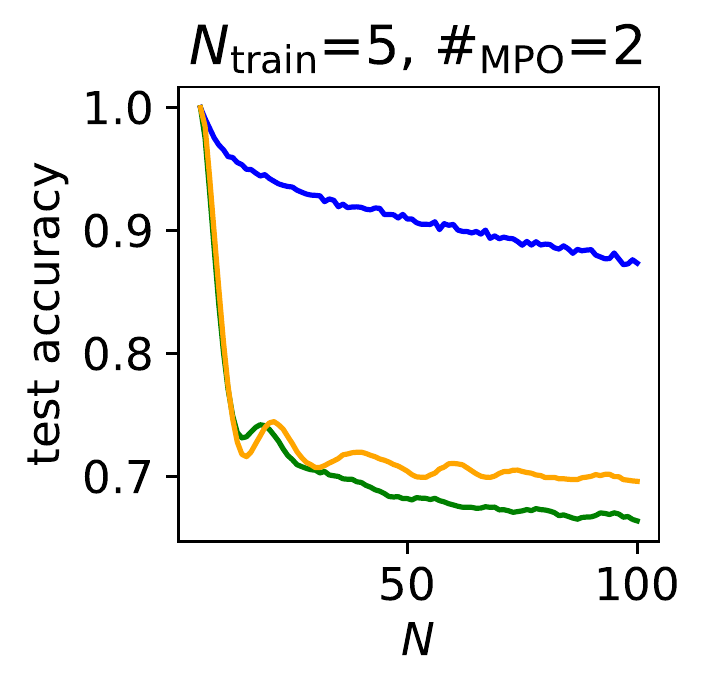}
    \includegraphics[width=0.245\textwidth]{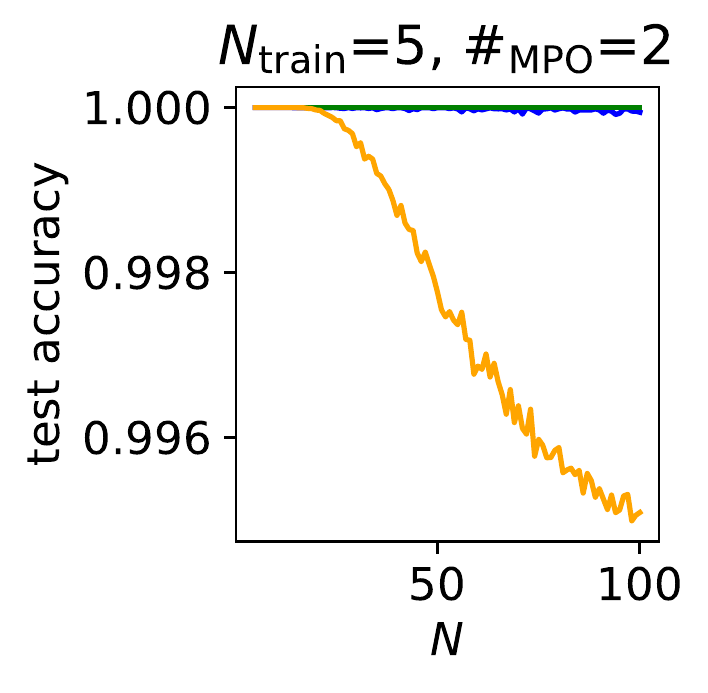}
    \includegraphics[width=0.245\textwidth]{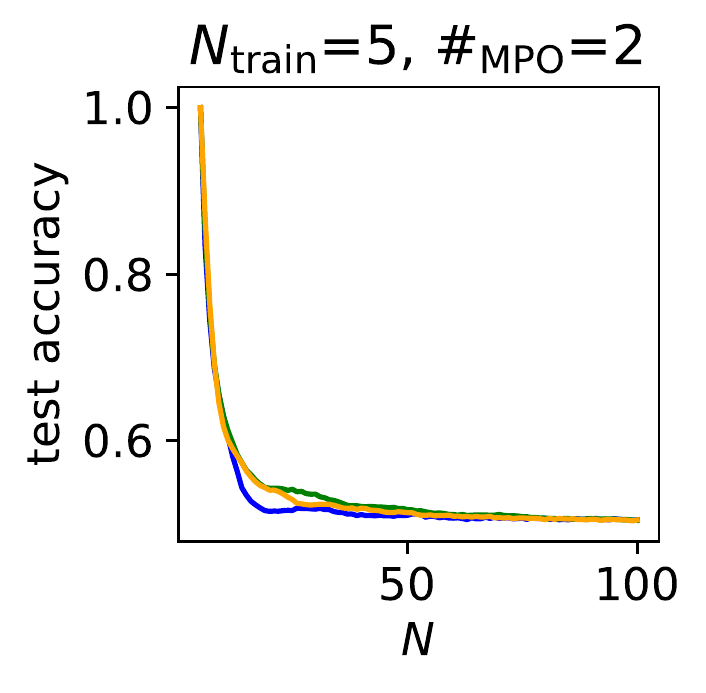}
    \includegraphics[width=0.245\textwidth]{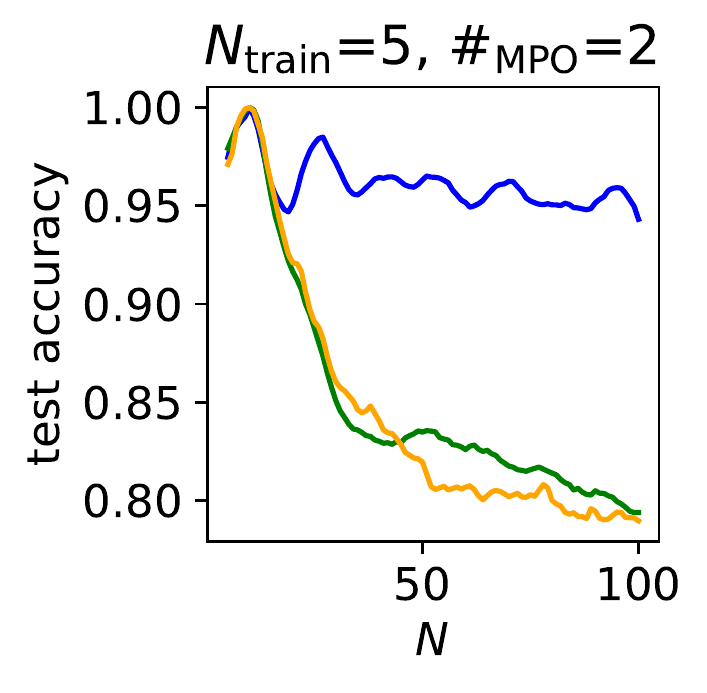}

    \includegraphics[width=0.4\textwidth]{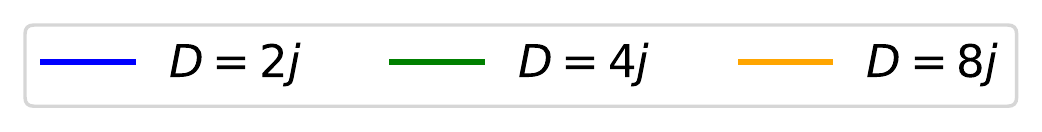}
    \caption{Generalisation of the fixed $N_{\rm training}=5, 10$ solutions of the $j=1$ (top row) and $j=2$ (bottom row) rule-30 problems to different input sizes $N=5,6\ldots 100$. We plot the average test accuracy for all runs that solve the problems for the fixed training input sizes $N_{\rm train}$.}
    
    \label{fig:ca generalisation}
\end{figure}

Considering the simplest $j=1,2$ tasks, we observed a remarkable generalisation. A solution for the fixed input size $N=5$ can generalise to all input sizes up to $N=100$. However, this is not always the case. In \fref{fig:ca generalisation} we show the average test error for test input sizes $N=5,6,\ldots 100$. We compare solutions with two different training input sizes $N_{\rm training}=5,10$ and models with one MPO layer ($\#_{\rm MPO}=1$) and two MPO layers ($\#_{\rm MPO}=2$). 

The generalisation performance is better for larger training input sizes and smaller $j$. In contrast to the image classification task, we find a better generalisation for models with only one layer. We can improve the generalisation and the training speed by training with a range of input sizes.

Finally, we numerically verify the exponential separation between deep and shallow models. For a fixed $j$, we found the minimal bond dimension $D_{\rm MPO}$, which solves the $j$-step problem. We considered a one-layer network ($D^{\rm 1-layer}_{\rm MPO}$) and a j-layer network ($D^{j{\rm-layer}}_{\rm MPO}$). As expected, the necessary bond dimension for the one layer network increases exponentially (see Table \ref{tab:D vs j fixed n}). On the other hand, we observed a slow growth of the bond dimension for the $j-$layer network. However, we found that the $j$-layer network is more difficult to train. We observe frequent sudden drop of accuracy. Also, we need more training runs to get the best solution. For example, we could not find the solution for the $4$-step problem with a 4-layer network. Therefore, we present the result for a 2-layer network instead. 
\begin{table}[!htb]
    \centering
    \begin{tabular}{c|c|c}
         $j$ & $D_{\rm MPO}^{\rm 1-layer}$ & $D_{\rm MPO}^{j{\rm-layers}}$ \\
         \hline
          1  & 2  & 2 \\
          2  & 4  & 2 \\ 
          3  & 8  & 3 \\
          4  & 18 & $8$ \textit{(2 layers)} \\ 
    \end{tabular}
    \caption{Minimal bond dimension for solving the $j-$step problem with one- and $j-$layer networks. The deep network for $j=4$ has only two MPO layers.}
    \label{tab:D vs j fixed n}
\end{table}

\section{Conclusions}
\label{sec:conclusion}
We introduced an MPO layer based on the time-dependent variational principle on matrix product states. We showed that the MPO layer is related to the linear dot-attention mechanism. It generalises it by taking into account also higher-order correlations. Therefore, it would be interesting to check how it can complement the standard attention mechanism used in the ubiquitous transformer architecture \cite{vaswani2017attention}. Another interesting application of the MPO transformation can be contextualisation of embeddings.

We evaluated the network on the image classification and the sequence prediction task. In the image classification case, we first introduced new baselines that already improved the current state-of-the-art for tensor network methods on the MNIS and FashionMNIST datasets. The baseline results on permuted datasets also suggest that the mutual information might not be a relevant indicator for the performance of the tensor network models on image classification tasks. Moreover, it is likely, that the augmentation procedure changes the nature of the correlations \cite{huang2017provably, lu2021tensor} and should be considered in such investigations. Deep tensor networks improve our baseline results even further and almost close the gap to baseline neural networks such as \cite{szegedy2015going} on the MNIST and Fashion MNIST datasets. We also showed that random crop training improves the robustness of uniform tensor network models to image size and aspect ratio changes. 

As an example of sequence modelling, we considered cellular automata. While deeper models need exponentially fewer parameters for the $j$-step predictions, they become increasingly hard to train. There seems to be a tradeoff between expressivity in terms of model parameters and training efficiency/stability. Finally, optimising the training procedures for deep tensor networks will probably lead to more robust training and improved results.

\section*{Acknowledgments}
BZ acknowledges support from Sloveinan research agency (ARRS) project J1-2480. Computational resources were provided by SLING – Slovenian national supercomputing network.

\section*{Declarations}
\textbf{Conflict of interest:} The authors declare no competing interests.

\bibliographystyle{unsrt}
\bibliography{deeptn}

\appendix
\section{Linear attention MPO}
\label{app:permutation mpo}
Here we show that we can rewrite a linear dot-attention as an MPO layer with the bond dimension $D_{\rm MPO} = d^2+1$. In particular, we show that a slight modification of the proposed MPO layer implements the linear dot-attention transformation
\begin{align}
\psi_j=(q_j\cdot k_l)q_l,
\label{eq:simple attention}
\end{align}
where $q_j=W^{\rm Q}\phi(j)$, and $k_j=W^{\rm K}\phi(j)$. For simplicity we removed the factor $\frac{1}{\sqrt{d}}$ and used a trivial transformation of values ($v_j=\phi(j)$). We obtain a more general transformation by applying the linear transformation $W^{\rm V}(W^{\rm Q})^{-1}$ on the transformed values $\psi_j$. Additionally, we assume that the embedding vectors are $L_2$ normalised.

The embedding tensor can be interpreted as an $N$-fold tensor product of embedding vectors $\Phi=\phi(1)\otimes\phi(2)\otimes\ldots\otimes\phi(N)$. We consider the action of a MPO on the vector $\Phi$. First, we decompose MPO tensors into three components
\begin{align}
M^{t,s}_{a,a'}(j)=\sum_{t',s'=1}^{d}W^{\rm Q}_{t',t}\tilde{M}^{t',s'}_{a,a'}(j)W^{\rm K}_{s',s}
\end{align}
The matrices $W^{\rm Q}$ and $W^{\rm K}$ transform the local vectors $\phi(j)$ to queries and keys. The remaining MPO implements a permutation operator  
\begin{align}
    {\mathrm Tr}\tilde{G} \tilde{M}^{t_1,s_1}\tilde{M}^{t_2,s_2}\ldots\tilde{M}^{t_N,s_N} = \sum_{i<j=1}^N P_{ij},
\end{align}
where
\begin{align}
    P_{i,j}\psi(1)\otimes\psi(2)\otimes\ldots\psi(i)\otimes\ldots\psi(j)\otimes\ldots\psi(N) = \psi(1)\otimes\psi(2)\otimes\ldots\psi(j)\otimes\ldots\psi(i)\otimes\ldots\psi(N) 
\end{align}
The local weight matrix, without the normalisation of the left and right context matrices, is then given by
\begin{align}
    H(j) = \sum_{i,l\neq j}(q_i\cdot k_l)(k_i\cdot q_l) \mathds{1}_d + \sum_{i\neq j} q_i k_i^T.
\end{align}
The final transformation of the embedding $\phi(j)$ with the described MPO layer is then
\begin{align}
    \phi(j)\rightarrow c q_j + \sum_{i\neq j}q_i (k_i\cdot q_j),
    \label{eq:mpo attention}
\end{align}
where $c=\sum_{i,l\neq j}(q_i\cdot k_l)(k_i\cdot q_l)$. We can correct the difference between the linear attention and the result \eref{eq:mpo attention} by a simple local residual connection. The final difference is that we normalise the left and the right context matrices. This normalisation translates to a rescaling of the final output and has no effect if we normalise the output after each application of the attention mechanism.

Finally, we provide the MPO tensors $\tilde{M}$ that implement the permutation transformation
\begin{align}
    \tilde{M}^{t,s}_{1,dt+s+1}&=1,\quad t,s=1,2\ldots d,\\
    \tilde{M}^{s,t}_{dt+s+1,1}&=1,\quad t,s=1,2\ldots d,\\
    \tilde{M}^{s,s}_{a,a}&=1,\quad s=1,2\ldots d,\quad a=2,3\ldots d^2+1,\\
    \tilde{G}_{1,1}&=1.
\end{align}
The remaining elements of the tensors $\tilde{M}$ and $\tilde{G}$ are zero.

\section{Image augmentation details}
\label{app:hyperparameters}
We use \textit{torchvision.transforms} to augment the datasets. We performed hyperparameter search for the given model to improve the various transformation parameters and probabilities. In particular, we used the following sequence of transformations ( Transformation-name(parameter 1, parameter 2,... parameter $n$). Along with parameters in the brackets, we also tuned the probability of applying the change):

\begin{itemize}
    
    \item Random pixel shift in the range [-aug\_phi, aug\_phi]
    
    \item Random color jitter: transforms.ColorJitter(brightness=brightness,  contrast=contrast, saturation=saturation, hue=hue)
    
    \item Random sharpness: transforms.functional.adjust\_sharpness(sharpness\_factor=sharp\_fac)
    
    \item Random Gaussian blur: transforms.GaussianBlur(kernel\_size=blur\_kernel\_size)
    
    \item Random horizontal flip: transforms.RandomHorizontalFlip()
    
    \item Random horizontal flip: transforms.RandomHorizontalFlip()
    
    \item Random affine transformation: transforms.RandomAffine(rotate, translate=(txy, txy), scale=(scale\_min, scale\_max))
    
    \item Random perspective transformation: transforms.RandomPerspective(distortion\_scale=perspective\_scale)
    
    \item Resize: this transformation is always applied transforms.Resize(resize)
    
    \item Random crop: transforms.RandomCrop(crop)
    
    \item Random elastic transformation: (here we used the elasticdeform library as import elasticdeform.torch as etorch ) etorch.deform\_grid(displacement)
    
    \item Random erasing: transforms.RandomErasing(scale=(erasing\_scale\_min, erasing\_scale\_max), ratio=(0.3, 3.3))
\end{itemize}

The order of transformation is fixed and is the same as given above, independent of the dataset. During hyperparameter search, the training was restricted to a maximum of 30 epochs.

The best parameter set was chosen based on the validation accuracy of the 5-fold cross-validation average. If several parameter configurations had the same average, we chose the one with the smallest average train accuracy. We expect these cases to have better generalization properties. We list the best parameter configurations in table \ref{tab:parameters}.

\begin{table}[!htb]
    \centering
    \begin{tabular}{c|c|c}
    Parameter & MNIST & FashionMNIST  \\
    \hline
    lr & 0.00026 & 8.1e-05 \\
    l2 & 0.0033 & 3.77e-06\\
    optimizer & adamw & adam\\
    step & 20 & 20\\
    gamma & 0.5 & 0.5\\
    crop & 25 & 28\\
    aug\_phi & 1.3e-05 & 0.00072\\
    aug\_color\_jitter\_prob & 0.69 & 0.81\\
    aug\_brightness & 0.49 & 0.11\\
    aug\_contrast & 0.11 & 0.46\\
    aug\_saturation & 0.22 & 0.23\\
    aug\_hue & 0.12 & 0.5\\
    aug\_sharpness\_prob & 0.21 & 0.59\\
    aug\_sharp\_min & 0.28 & 0.42\\
    aug\_sharp\_max & 1.56 & 5.45\\
    aug\_gblur\_prob & 0.35 & 0.31\\
    aug\_gblur\_kernel & 2 & 19\\
    aug\_horizontal\_filp & 0 & 0\\
    aug\_affine\_prob & 0.0064 & 0.15\\
    aug\_translate & 0.10 & 0.11\\
    aug\_rotate & 0.93 & 11.35\\
    aug\_scale\_min & 0.85 & 0.85\\
    aug\_scale\_max & 1.01 & 1.1\\
    aug\_perspective\_prob & 0.34 & 0.67\\
    aug\_perspective\_scale & 0.49 & 0.14\\
    aug\_elastic\_prob & 0.63 & 0.24\\
    aug\_elastic\_strength & 0.27 & 3.31\\
    aug\_erasing\_prob & 0.34 & 0.13\\
    aug\_erasing\_scale\_min & 0.018 & 0.0086\\
    aug\_erasing\_scale\_max & 0.091 & 0.3
    \end{tabular}
    \caption{Best hyperparameter settings after an extensive 5-fold cross-validation hyperparameter search for the MNIST and FashionMNIST datasets.}
    \label{tab:parameters}
\end{table}


\end{document}